\documentclass[pmlr]{jmlr}

\usepackage{soul}
\usepackage[utf8]{inputenc}
\usepackage{CJKutf8}
\usepackage{array}
\usepackage{ragged2e}
\usepackage{booktabs}
\usepackage{float}
\usepackage{algorithmic}
\usepackage{tikz}
\usetikzlibrary{fit,tikzmark}

\hypersetup{hypertexnames=false}
\newcommand{\subcaptiontext}[1]{\par\small #1\par}

\jmlrproceedings{PMLR}{Proceedings of Machine Learning Research}
\jmlrvolume{340}
\jmlryear{2026}
\jmlrworkshop{Machine Learning for Healthcare}

\title[Key Coverage Matters]{Key Coverage Matters: Semi-Structured Extraction of \\OCR Clinical Reports}

\author{
\Name{Yu Wang}
\addr AI Starfish\\
Hangzhou, China
\AND
\Name{Yingyun Li}
\addr AI Starfish\\
Hangzhou, China
\AND
\Name{Ying Qin}
\addr AI Starfish\\
Hangzhou, China
\AND
\Name{Haiyang Qian}
\Email{haiyang.qian@aistarfish.com}
\addr AI Starfish\\
Hangzhou, China}

\begin{document}
\begin{CJK}{UTF8}{gbsn}
\maketitle

\begin{abstract}
This work addresses a cross-institution workflow in which patients present paper or scanned reports from prior visits because relevant records are not available in the receiving hospital's EHR. This hinders not only electronic health record (EHR) integration and longitudinal review, but also downstream workflows that depend on more complete patient records, including cross-institution EHR back-fill, longitudinal follow-up for chronic disease and oncology, and report-derived clinical-trial eligibility screening. Although optical character recognition (OCR) can digitize such reports, reliable extraction remains challenging because clinical documents are heterogeneous, OCR text is noisy, and many healthcare settings require low-cost on-premise deployment.

We formulate this problem as canonical key-conditioned extractive question answering and maintain a canonical inventory and alias mapping through iterative key mining, normalization, clustering, and incremental human verification. On the 849-report development set, inventory expansion increases end-to-end Exact recall from 0.3534 to 0.8042, with Exact F1 peaking at 0.8232 for Top-95. A supplementary gold-key-conditioned comparison shows that a compact 0.2B BERT--BiLSTM--CRF extractor runs substantially faster than LoRA-adapted Qwen3 models on the same GPU, with a modest trade-off in Exact F1. An independent downstream study found task-dependent effects of structured fields, including gains of 3.27 percentage points for histologic grade and 3.52 points for treatment response, but the effects were not uniformly positive across tasks; this study evaluates representation utility rather than the proposed extractor. Although the corpus is Chinese, the method is based on the language-agnostic key--value organization of semi-structured clinical reports and can be adapted to other settings with an appropriate canonical inventory and alias mapping.
\end{abstract}


\section{Introduction}
\begin{figure}[ht]
    \centering
    \includegraphics[width=\columnwidth]{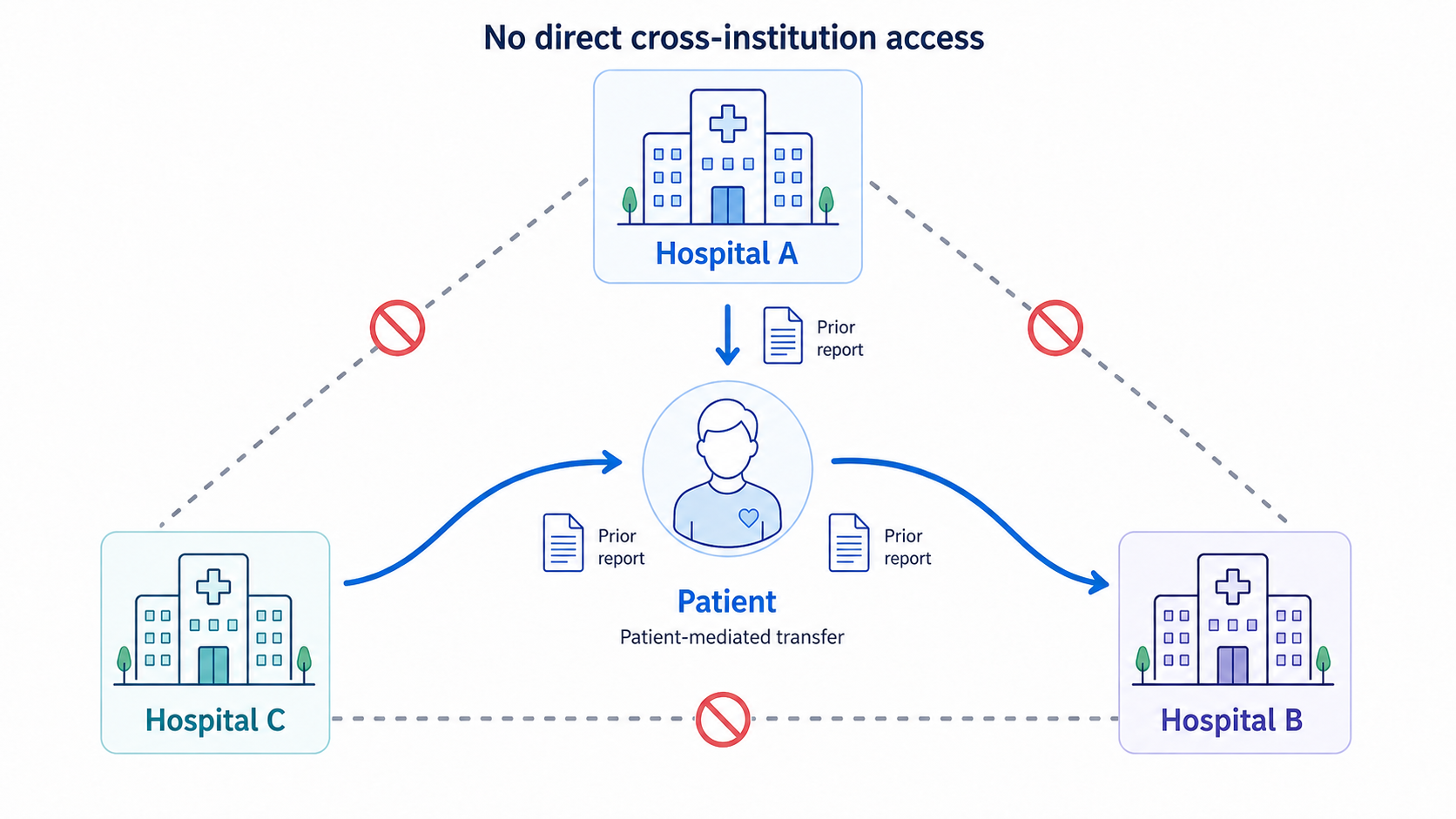}
    \caption{Cross-institution data silos in clinical document processing pipelines.}
    \label{fig:data-silos}
\end{figure}

Privacy constraints and institutional data silos can leave prior records outside the receiving hospital's EHR. The workflow considered in this study arises when patients present paper or scanned reports from previous visits to convey relevant medical history. Figure~\ref{fig:data-silos} depicts this scenario.
This practice poses significant challenges for both clinicians and patients. For clinicians, paper reports are inconvenient to read and cannot be integrated into EHR systems, limiting their usability for longitudinal review, structured retrieval \cite{hossain2023nlp_ehr}, and clinical decision support. Even after applying OCR, the resulting text often remains difficult to interpret due to recognition errors, layout noise, and inconsistent formatting \cite{li2024improving}, and still lacks native compatibility with EHR schemas. From the patient perspective, the lack of a unified and structured electronic record makes it difficult to maintain a coherent view of medical history. This fragmentation hinders continuity of care, particularly for patients with complex or chronic diseases such as cancer. Together, these challenges highlight the need for systematic methods that can transform heterogeneous, OCR-derived clinical reports into structured, standardized representations suitable for EHR integration and continuity of patient care \cite{walker2023perspectives}. In practice, such an extraction layer can back-fill missing fields from outside reports into a receiving institution's EHR, assemble longitudinal timelines for chronic-disease or oncology follow-up, and expose report-derived criteria for clinical-trial eligibility screening. These are intended workflow uses, not claims that this study improves clinical decisions, trial enrolment, or patient outcomes.

\begin{figure}[ht]
    \centering
    \begin{minipage}[t]{0.48\textwidth}
        \centering
        \includegraphics[width=\linewidth]{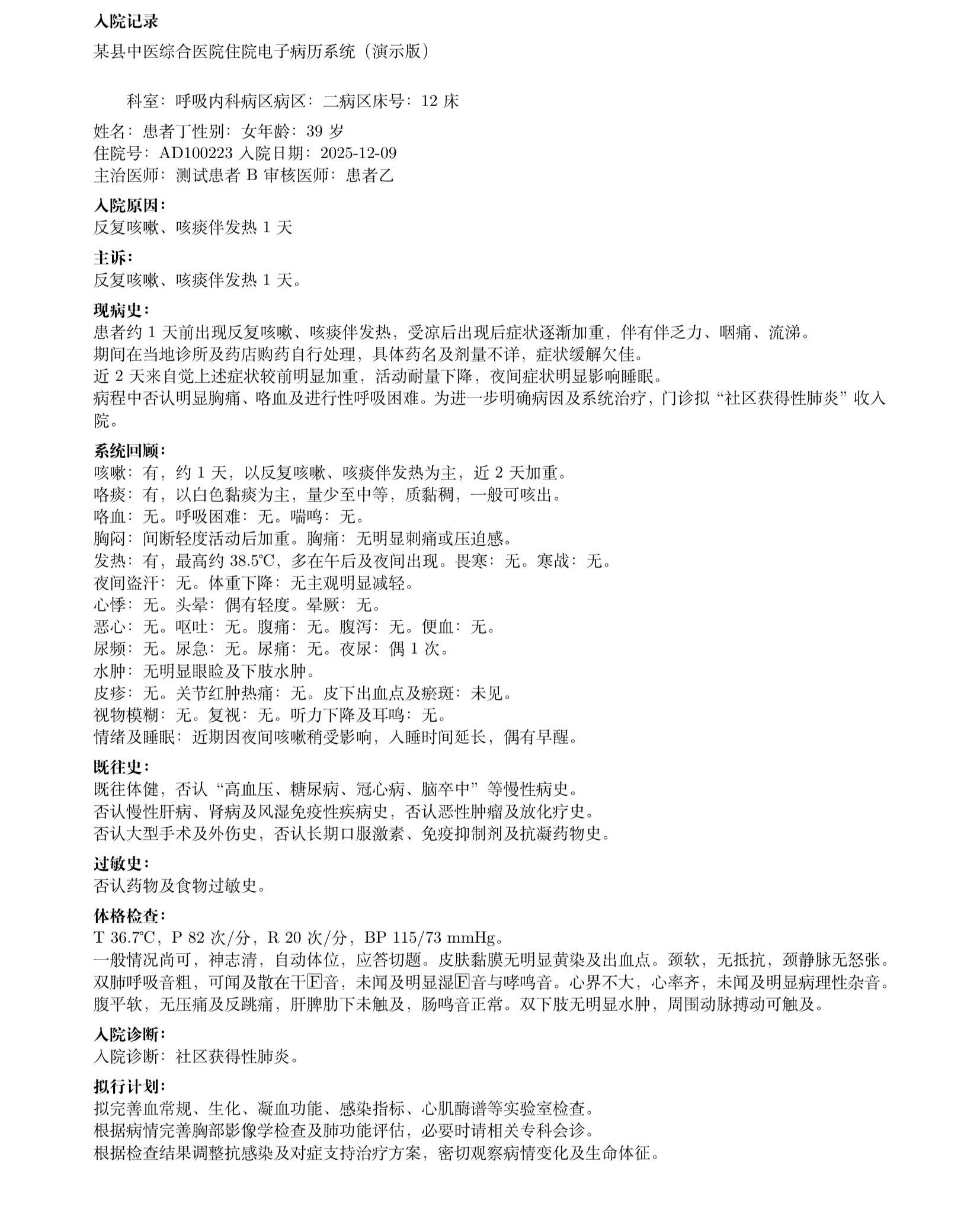}
        \subcaptiontext{(a) The admission record in Chinese}
    \end{minipage}
    \hfill
    \begin{minipage}[t]{0.48\textwidth}
    \centering
    \includegraphics[width=\linewidth]{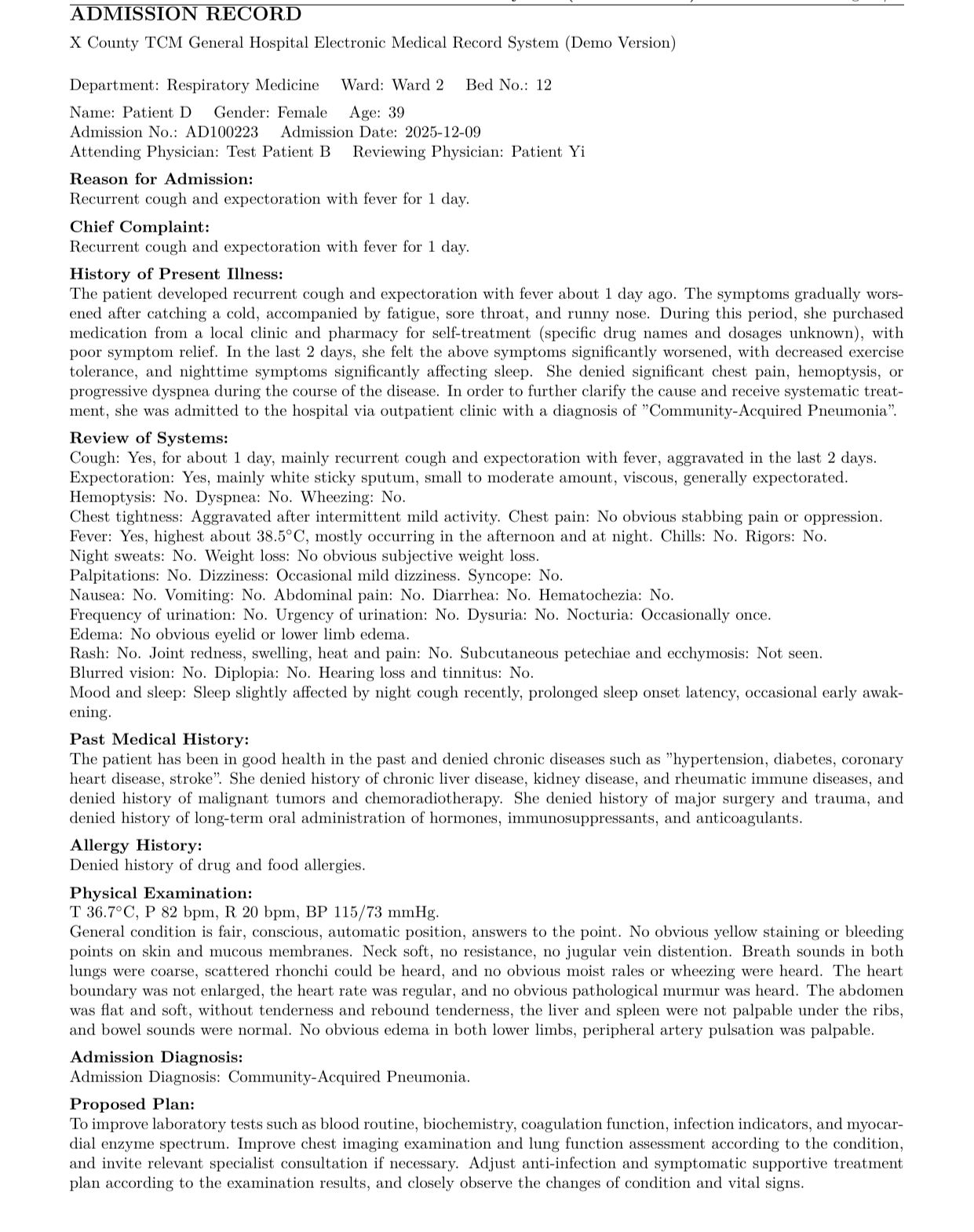}
    \subcaptiontext{(b) The English translation}
    \end{minipage}
    \caption{A sample admission report and its English translation.}
    \label{fig:clinical-report-example}
\end{figure}

Clinical reports often follow an explicit layout: clinical concepts appear as field headers (keys) followed by free-text content (values), separated by delimiters such as colons or layout cues \cite{fu2020clinical_concept_extraction,jaume2019funsd}. Figure~\ref{fig:clinical-report-example} illustrates a typical patient admission report. This key--value structure is common across languages and across mixed-form clinical documents. Owing to wording differences across institutions, the same clinical concept may appear under different surface-form keys, yielding a highly variable and long-tailed key space in which previously unseen keys continue to emerge \cite{wu2020deep_learning_clinical_nlp}.

Table~\ref{tab:surface-form} shows examples of institution-specific surface-form variations for the same clinical keys in Chinese, together with their English translations. Even when the underlying clinical meaning is the same, different hospitals may use different section headers or expressions.
\begin{table}[ht]
\centering
\small
\begin{tabular}{
>{\RaggedRight\arraybackslash}p{0.10\textwidth}
>{\RaggedRight\arraybackslash}p{0.25\textwidth}
>{\RaggedRight\arraybackslash}p{0.15\textwidth}
>{\RaggedRight\arraybackslash}p{0.30\textwidth}
}
\toprule
\multicolumn{2}{c}{\textbf{Chinese}} & \multicolumn{2}{c}{\textbf{English}} \\
\cmidrule(lr){1-2} \cmidrule(lr){3-4}
\textbf{Key} & \textbf{Surface Forms} & \textbf{Key} & \textbf{Surface Forms} \\
\midrule

既往史 &
既往病史，既往疾病，既往治疗史 &
Past Medical History &
Past Illness History, Previous Diseases, Previous Treatment History \\
\addlinespace

体格检查 &
查体所见，体检所见，体征情况 &
Physical Examination &
Physical findings, Examination Findings, Clinical Signs \\
\addlinespace

手术记录 &
手术经过，术中经过，手术过程，术中情况 &
Operative Record &
Surgical Course, Intraoperative Course, Surgical Procedure, Intraoperative situation \\
\addlinespace

专科检查 &
专科查体，专科检查所见，专科情况，专科查体所见 &
Specialty Examination &
Specialty Physical Examination; Specialty Examination Findings; Specialty Status; Specialty Physical Examination Findings \\
\bottomrule
\end{tabular}
\caption{Examples of surface-form keys in Chinese clinical reports (and English translations).}
\label{tab:surface-form}
\end{table}

\begin{figure}[ht]
    \centering
    \includegraphics[width=\columnwidth]{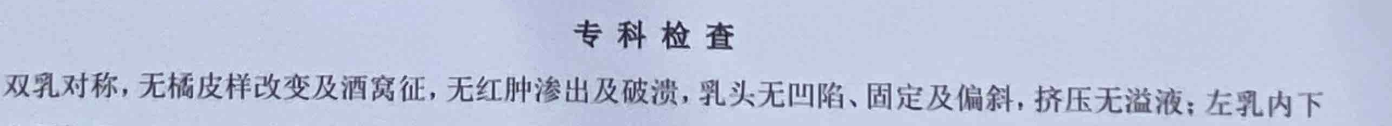}
    \subcaptiontext{(a) Specialty Examination (专科检查)}
    \vspace{2pt}

    \includegraphics[width=\columnwidth]{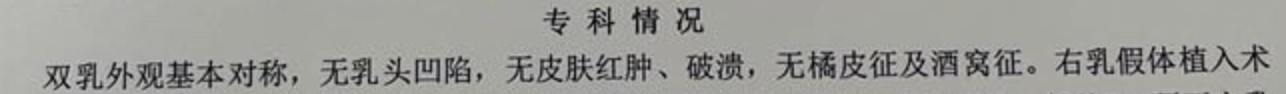}
    \subcaptiontext{(b) Specialty Status (专科情况)}
    \vspace{2pt}

    \includegraphics[width=\columnwidth]{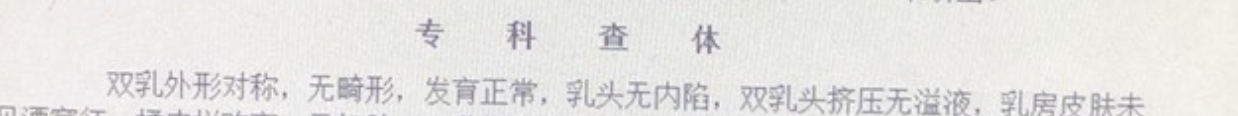}
    \subcaptiontext{(c) Specialty Physical Examination (专科查体)}

    \caption{Examples of surface-form variants mapped to the canonical key ``Specialty Examination''.}
    \label{fig:realReport}
\end{figure}

Figure~\ref{fig:realReport} provides a real-world example of such variation for the key ``Specialty Examination''. The three report snippets use different surface forms, all of which map to the canonical key ``Specialty Examination''.
 
Treating field headers as queries and their contents as answers yields an extractive question answering (QA) formulation, but standard QA-based extractors assume a predefined, closed key set. Clinical report keys instead form an open and evolving space: surface forms vary across institutions and templates, and new forms continue to appear. Thus, inventory coverage limits the range of fields that a key-conditioned system can recover.

We address this problem with a canonical-key-conditioned extraction framework for semi-structured OCR clinical reports. The current inventory determines which fields are queried, so inventory-external fields cannot enter the accepted output; normalization and iterative clustering maintain canonical keys and their aliases as new surface forms are reviewed. Our central experiment evaluates train-derived Top-$N$ inventories and their corresponding models under a common architecture and fixed development set. By reporting coverage, conditional extraction, and end-to-end recovery separately, we distinguish the supported field range from value extraction once a field is supported. Gold-key-conditioned accuracy and efficiency comparisons therefore provide conditional rather than inventory-coverage evidence.

\subsection*{Generalizable Insights about Machine Learning in the Context of Healthcare}
Canonical-key coverage is an explicit constraint on end-to-end field recovery. High conditional extraction accuracy cannot substitute for evaluating inventory coverage: when the canonical inventory does not represent valid surface forms or clinically relevant concepts, information present in the original report cannot enter the structured data. This can affect patient cohort construction, research analyses, and reuse of clinical information. Separating inventory coverage from conditional extraction performance therefore helps identify whether errors arise from the representation layer or the extraction model. This evaluation principle can also apply to other semi-structured document tasks in which heterogeneous surface forms must be mapped to a shared schema.

\section{Related Work}

\paragraph{Clinical document and OCR extraction.}
Transferred clinical records are often scanned or printed images that require OCR before their contents can be structured. OCR-derived clinical text contains character and layout noise that can impair downstream extraction~\cite{li2024improving}. MedStruct-S extends this setting to semi-structured clinical reports with open-world keys and OCR noise~\cite{li2026medstructs}, closely matching our focus on robust page-level span extraction.

\paragraph{Query-based field extraction.}
Question answering provides a practical formulation for extracting fields from clinical records with variable layouts. EMRQA established this paradigm for medical record extraction~\cite{pampari2018emrqa}, and subsequent work has applied it to radiology reports~\cite{dada2024information}. Biomedical encoders including BioBERT~\cite{lee2020biobert} and PubMedBERT~\cite{gu2021pubmedbert} support both QA and information-extraction settings.

\paragraph{Key normalization and open-world extraction.}
Semi-structured report keys vary across institutions and over time. This resembles biomedical entity normalization, which maps surface names to underlying concepts~\cite{liu2021self}, but here the normalized objects are report keys. In contrast to extraction methods that assume a fixed, known key set, we quantify the effect of incomplete inventories and use the result to guide inventory maintenance and expansion~\cite{lu-etal-2022-unified,shen-etal-2022-parallel}.

\section{Methodology}
In our method, we treat each clinical-report page as the atomic unit of extraction. If a report contains multiple pages, we aggregate the page-level outputs.
\subsection{Dataset and Annotation}
We analyze a private corpus of oncology clinical-report images collected through a multi-institution oncology patient-management program, reflecting heterogeneity in clinical documentation and report templates. The corpus covers a diverse set of report categories, including outpatient medical records, biopsy pathology reports, examination records, postoperative pathology reports, genetic testing reports, admission records, and consultation pathology reports. Among these, outpatient medical records are the most frequent category, accounting for approximately 35\% of all reports.

All report images were digitized on premises with a privately deployed Baidu OCR V6 service~\citep{baidu_ocr}; no clinical image was sent to a public OCR endpoint. The extraction models consume only the page-level character sequence formed by concatenating recognized line strings in the engine-returned reading order; line/character confidence and geometry are stored as metadata and are not used as model inputs. Request parameters and response-normalization details are given in Appendix~\ref{app:ocr-config}.

To protect patient privacy, all personally identifiable information—including names, identification numbers, contact information, and addresses—was removed prior to annotation and replaced with a uniform placeholder symbol (\texttt{**}). Only de-identified OCR text was used in subsequent processing and analysis.

The corpus was annotated at the surface-form key--value (KV) pair level through a human-in-the-loop workflow. A locally deployed Qwen3-235B-A22B model~\citep{qwen3} generated initial KV pre-annotations, which human reviewers corrected in Label Studio~\citep{labelstudio}; unresolved cases were escalated for senior adjudication. The available records identify 29 distinct non-staff annotation accounts, but do not permit reliable inference about the number or clinical training of individual annotators. The recorded 430 human-hours refer to review and correction rather than annotation from scratch, corresponding to approximately 1.6 minutes per report or 11.7 seconds per KV pair. These are two normalizations of the same aggregate review time.

After exact full-OCR-text deduplication, 14,587 tasks were partitioned into 13,128 training tasks and 1,459 held-out tasks. These held-out tasks support the supplementary conditional-extraction and robustness analyses, whose shared evaluation set contains the 1,242 reports with valid reference spans (Appendix~\ref{app:corpus-flow}). The Chinese coverage experiment instead used a fixed 7,639/849 train/development split from a distinct 8,488-report subset.

\begin{figure}[H]
    \centering

    \begin{minipage}[t]{0.48\textwidth}
        \centering
        \includegraphics[width=\linewidth]{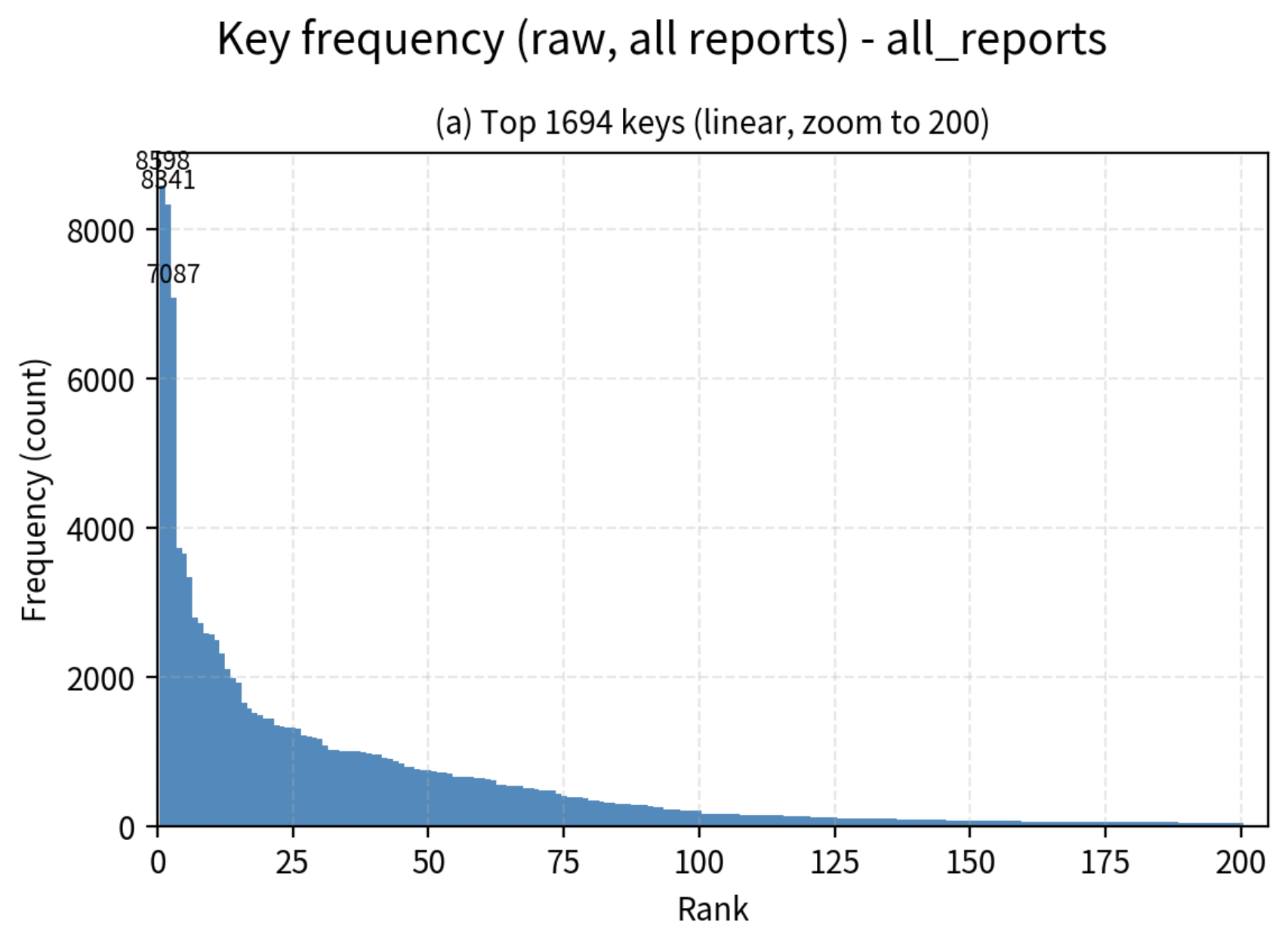}
        \subcaptiontext{(a) Rank--frequency plot (top-200, linear scale)}
    \end{minipage}
    \hfill
    \begin{minipage}[t]{0.48\textwidth}
        \centering
        \includegraphics[width=\linewidth]{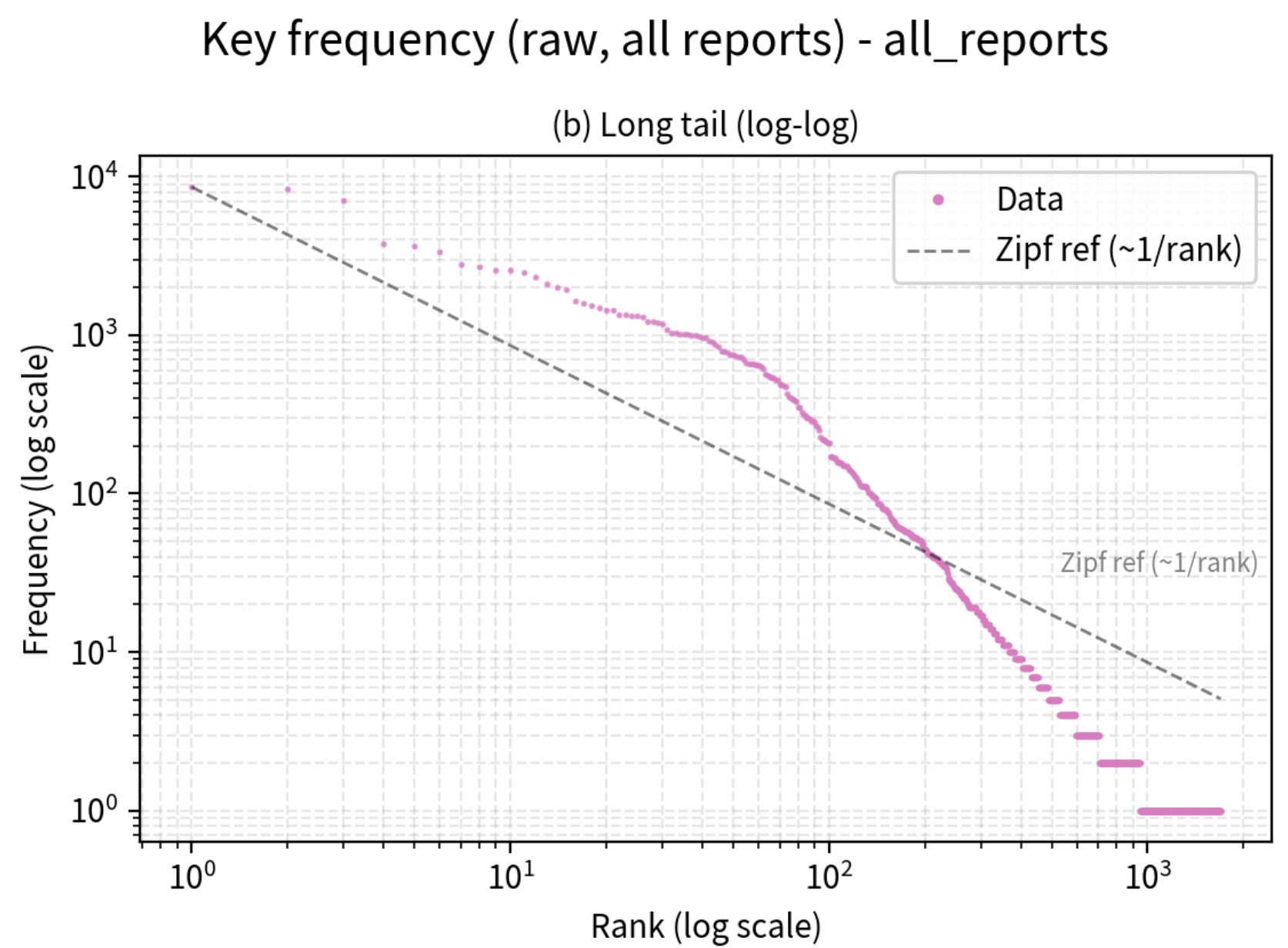}
        \subcaptiontext{(b) Log--log rank--frequency plot with a Zipf reference}
    \end{minipage}
    \caption{Rank--frequency characteristics of surface-form key expressions in the full corpus.}
    \label{fig:surface-form-frequency}
\end{figure}

The distribution of surface-form keys exhibits a strong long-tailed pattern (Figure~\ref{fig:surface-form-frequency}): a small set of frequent keys coexists with many rare surface forms. The denominator and split provenance required to report exact cumulative coverage thresholds have not been verified; we therefore omit those values.
\begin{figure}[t]
\centering
\refstepcounter{algocf}
\small
\textbf{Algorithm~\thealgocf: Iterative Canonical Key Expansion and Key--Value Extraction}
\label{alg:iterative-key-expansion}
\begin{minipage}{0.96\linewidth}
\begin{algorithmic}[1]

\REQUIRE
Initial canonical key set $\mathcal{K}_{\mathrm{c}}^{(0)}$,
initial alias mapping $\mathcal{K}_{\mathrm{a}}^{(0)}$,
clinical report batches $\{\mathcal{D}_b\}_{b=1}^{T}$

\ENSURE
Canonicalized key--value pairs $\mathcal{S}_{\mathrm{c}}$

\STATE Initialize QA models $\mathrm{QA}_{\mathrm{v}}$ and $\mathrm{QA}_{\mathrm{k}}$ using augmented data
\STATE $\mathcal{K}_{\mathrm{c}} \leftarrow \mathcal{K}_{\mathrm{c}}^{(0)}$,
$\mathcal{K}_{\mathrm{a}} \leftarrow \mathcal{K}_{\mathrm{a}}^{(0)}$

\FOR{$b = 1$ \TO $T$}
    \STATE Observe new report batch $\mathcal{D}_b$
    \STATE Initialize observed key set $\mathcal{K}(\mathcal{D}_b) \leftarrow \emptyset$

    \FOR{each report page with OCR text $\mathcal{T} \in \mathcal{D}_b$}
        \FOR{each canonical key $kc \in \mathcal{K}_{\mathrm{c}}$}
            \STATE Query value:
            $v_{kc} \leftarrow \mathrm{QA}_{\mathrm{v}}(q_{\mathrm{v}}(kc), \mathcal{T})$
            \STATE Query surface-form key:
            $k_{kc} \leftarrow \mathrm{QA}_{\mathrm{k}}(q_{\mathrm{k}}(kc), \mathcal{T})$
            \IF{$k_{kc} \neq \emptyset$}
                \STATE $\mathcal{K}(\mathcal{D}_b) \leftarrow \mathcal{K}(\mathcal{D}_b) \cup \{k_{kc}\}$
                \STATE Add $(k_{kc}, v_{kc})$ to $\mathcal{S}_{\mathrm{c}}$
            \ENDIF
        \ENDFOR
    \ENDFOR

    \STATE Identify novel surface-form keys:
    $\mathcal{K}_{\text{new}} \leftarrow \mathcal{K}(\mathcal{D}_b) \setminus \left( \mathcal{K}_{\mathrm{c}} \cup \bigcup_{kc \in \mathcal{K}_{\mathrm{c}}} \mathcal{K}_{\mathrm{a}}(kc) \right)$

    \STATE Format and clean $\mathcal{K}_{\text{new}}$
    \STATE Canonicalize $\mathcal{K}_{\text{new}}$ via embedding-based clustering with human verification
    \STATE Update canonical key set $\mathcal{K}_{\mathrm{c}}$ and alias mapping $\mathcal{K}_{\mathrm{a}}$
    \STATE Fine-tune QA models with expanded $\mathcal{K}_{\mathrm{c}}$ and $\mathcal{K}_{\mathrm{a}}$

\ENDFOR

\RETURN $\mathcal{S}_{\mathrm{c}}$

\end{algorithmic}
\end{minipage}
\end{figure}

\subsection{Problem Formulation}
We represent the OCR-derived text of a clinical report page as an ordered sequence of characters, denoted by
\begin{equation*}
\mathcal{T} = \big( c_1, c_2, \cdots, c_{M} \big), 
\end{equation*}
where $M$ is the total number of OCR characters in the page, and the order of characters preserves the original reading order of the document. 
Let $\mathcal{K}^\star$ denote the latent universe of all possible keys that may appear in real-world clinical reports and let $\mathcal{K}$ denote the set of all annotated keys collected during the labeling process. Note that $\mathcal{K}^\star$ cannot be fully observed in practice. That is, $\mathcal{K} \subset \mathcal{K}^\star$. The keys in $\mathcal{K}$ may include lexical variants or synonyms that correspond to the same underlying semantic meaning. We introduce a canonical key space:
\begin{align*}
    &\mathcal{K}_{\mathrm{c}} =  \{ {kc}_1, {kc}_2, \cdots, {kc}_\mathrm{K}\}, \\
    &\textrm{where } K\textrm{ is the number of observed canonical keys.}
\end{align*}
We define an alias set for each observed canonical key as: $\mathcal{K}_{\mathrm{a}}(kc) \subseteq \mathcal{K},$ where all elements in $\mathcal{K}_{\mathrm{a}}(kc)$ are semantically equivalent to $kc$,  and $kc$ is not in the set of $\mathcal{K}_\mathrm{a}(kc)$. The observed space $\mathcal{K}$ is the union of all alias sets and canonical sets:
\begin{equation*}
\mathcal{K} = \mathcal{K}_{\mathrm{c}}
\;\cup\;\bigcup_{kc \in \mathcal{K}_{\mathrm{c}}}
\mathcal{K}_{\mathrm{a}}(kc).
\end{equation*}
We define a key canonicalization function:
\begin{equation*}
\psi : \mathcal{K} \rightarrow \mathcal{K}_{\mathrm{c}},    
\end{equation*}
such that:
\begin{equation*}
\psi(k) = kc
\quad \text{if and only if} \quad
k \in \mathcal{K}_{\mathrm{a}}(kc) \cup \{kc\}.    
\end{equation*}
This mapping function collapses synonymous keys into a single canonical representation. During inventory expansion, newly discovered surface-form keys are formatted, semantically clustered, and reviewed before being added to the canonical inventory. Under the documented review protocol, annotators confirm proposed canonical merges, manually split a proposed cluster when it mixes distinct field semantics, and route unresolved assignments to senior review; the senior ruling governs the final canonical-key and alias decision.

Let $k \in \mathcal{K}$ denote a surface-form key observed in a report, $v$ denote its value, and $\mathcal{V}$ denote the value set that may appear in the real-world clinical reports. Thus, the key--value pair is represented by $(k,v)$. The canonical key--value pair is represented by:
\begin{equation*}
\big( \psi(k), v \big) \in \mathcal{K}_{\mathrm{c}} \times \mathcal{V}.
\end{equation*}
For the purpose of formulation, we first assume that an incomplete but known canonical key set is provided as prior knowledge. There are two steps in extracting the key--value pairs. First, we formulate value extraction as a QA task conditioned on canonical keys. The query denoted by $q_\mathrm{v}(kc)$ represents the question conditioned on a canonical key. A BERT-based QA model is then applied to the OCR token sequence: 
\begin{equation}
\label{eq:value_extraction}
\mathrm{QA}_{\mathrm{v}} :
\big( q_\mathrm{v}(kc), \mathcal{T} \big)
\;\mapsto\;
v_{kc} \in \mathcal{V} \cup \{ \emptyset \}.
\end{equation}
The output $v_{kc}$ denotes the extracted value corresponding to a canonical key or $\emptyset$ if no value is found. The output key--value space is given by:
\begin{equation*}
    \mathcal{S}_{\mathrm{c}}=\big\{
(kc, v_{kc})\;\big|\;kc \in \mathcal{K}_{\mathrm{c}},\ v_{kc} \neq \emptyset \big\}.
\end{equation*}
For any canonical key, $kc_i$, the simplest prompt for the QA task is:\\
\texttt{Extract the value of the key  $kc_i$}. \\
To help the model recognize synonymous surface forms, the query may include some or all known aliases of the canonical key:\\
\begin{quote}
\small\ttfamily\raggedright
Extract the value of the key $kc_i$. The key in the text may be a variant of
the canonical key, such as $ka_1, ka_2,\ldots$.
\end{quote}
where $ka_1, ka_2, \ldots \in \mathcal{K}_{\mathrm{a}}(kc_i)$.

Secondly, we identify the surface-form alias of each canonical key as it appears in the original OCR text. Using the same BERT-based QA model, we pose a second query for each canonical key: $q_{\mathrm{k}}(kc)$. Formally:
\begin{equation}
\label{eq:original_key_extraction}
    \mathrm{QA}_{\mathrm{k}} :\big( q_{\mathrm{k}}(kc), \mathcal{T} \big)\;\mapsto\; k_{kc} \in \mathcal{K} \cup \{ \emptyset \}.
\end{equation}
The output $k_{kc}$ denotes the original key observed on the page or $\emptyset$ if no alias is present. 

Note that the example prompt is illustrated in English. In our implementation, its equivalent Chinese version is adopted. Combining Eqs.~\eqref{eq:value_extraction} and \eqref{eq:original_key_extraction} yields $(k_{kc},v_{kc})$.

\begin{figure}[t]
    \centering
    \includegraphics[width=\columnwidth]{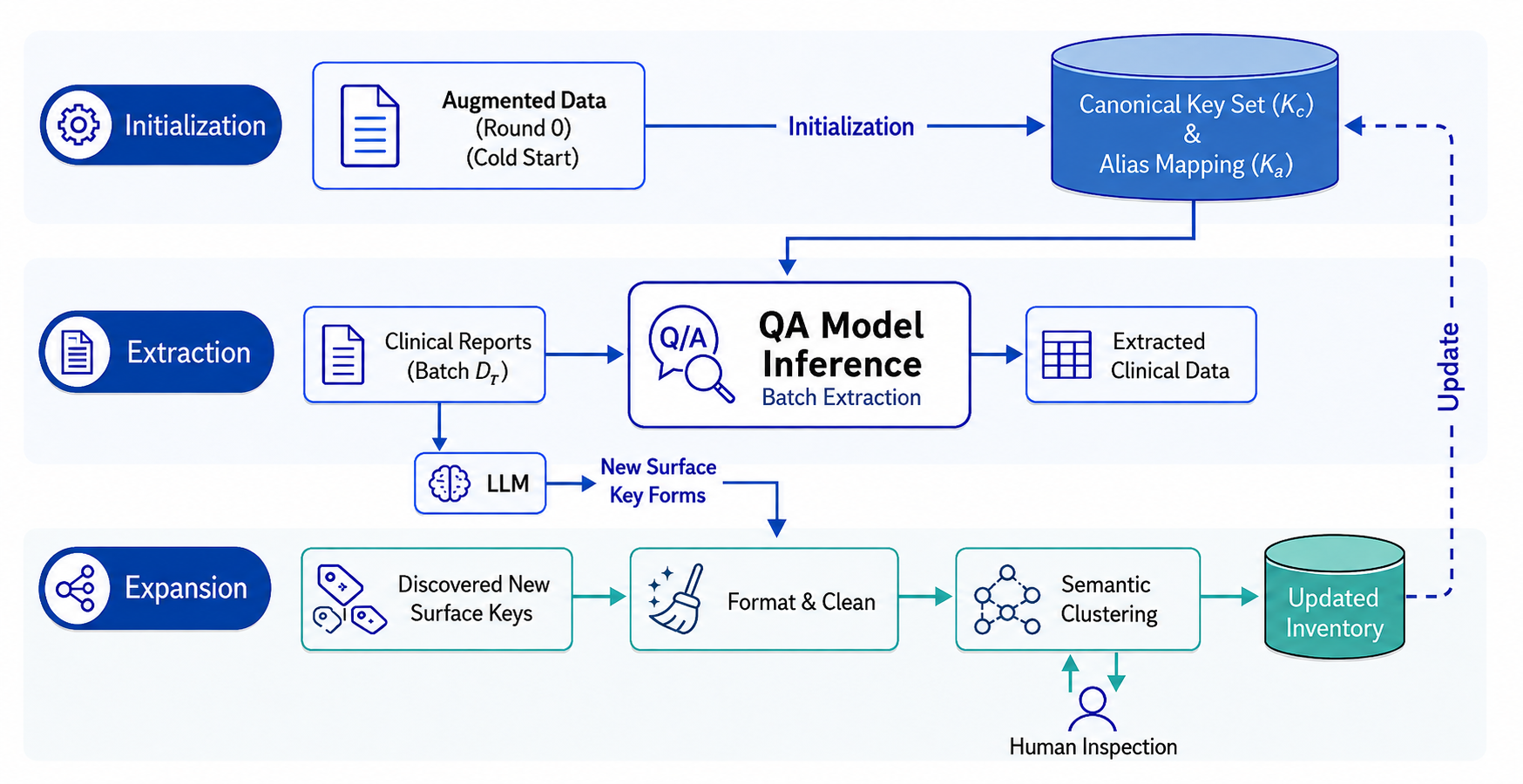}
\caption{Overview of the proposed iterative pipeline}
\label{fig:method_overview}
\end{figure}

In practice, neither the canonical inventory $\mathcal{K}_{\mathrm{c}}$ nor the alias mapping $\mathcal{K}_{\mathrm{a}}$ is available as prior knowledge. Given a report corpus that grows as successive batches $\{\mathcal{D}_b\}_{b=1}^{T}$, newly observed surface-form keys expand the observed key space over time, so key mining and canonicalization are treated as an iterative process rather than one-time preprocessing. In the expansion step of Fig.~\ref{fig:method_overview}, the LLM component is a locally deployed Qwen3-235B-A22B model~\citep{qwen3}, which generates candidate New Surface Key Forms from each incoming report batch. These candidates are formatted and cleaned, embedded with BGE-large-zh-v1.5, and grouped using average-linkage clustering with cosine distance threshold 0.25. The threshold is selected from a silhouette sweep, and the documented human-review rules above are applied before the canonical inventory and alias mapping are updated. BGE is used in this canonicalization stage, not for candidate surface-key identification or KV pre-annotation. The downstream QA models are then updated with the expanded inventory. With a document length $L$ and $K$ available canonical keys, inference requires $K$ key-conditioned queries and scales as $O(LK)$. This cost is the implementation trade-off that makes the inventory an explicit gate on supported fields.

\subsection{Evaluation Method}
Exact span matching is the primary metric. We additionally report BTM-3 to
distinguish small punctuation- or whitespace-boundary differences from larger
extraction errors. Before comparison, both strings undergo the same deterministic
normalization $\nu(\cdot)$: Unicode NFKC normalization, dash unification,
edge-whitespace removal, and removal of common punctuation at the two string
boundaries. Exact matching then requires equality after normalization between a
predicted value $\hat{v}$ and its gold value $v$:
\begin{equation*}
\mathrm{Exact}(\hat{v}, v) =
\begin{cases}
1, & \text{if } \nu(\hat{v}) = \nu(v), \\
0, & \text{otherwise}.
\end{cases}
\end{equation*}

The second criterion is BTM-3, a boundary-tolerant match that applies an
Edge-3 boundary rule. Let $\tau_3(\cdot)$ remove at most three characters in
total from the two boundaries when those characters are whitespace or
punctuation. We compute
\begin{equation*}
\mathrm{BTM\text{-}3}(\hat{v}, v) =
\begin{cases}
1, & \text{if } \nu(\tau_3(\hat{v})) = \nu(\tau_3(v)), \\
0, & \text{otherwise}.
\end{cases}
\end{equation*}
This criterion tolerates small boundary artifacts without granting partial
credit for an arbitrary interior mismatch. Text-overlap matching, defined by
substring containment after normalization, is reported only as a diagnostic.

Under both criteria, we report precision, recall, and F1 conditioned on canonical keys.

To assess overall system performance, we further compute precision, recall, and F1 over surface-form key--value pairs, where a prediction is considered correct only if both the surface-form key and its associated value are correctly extracted under the specified matching criterion.

Key coverage quantifies the proportion of gold field instances supported by the current canonical inventory. For a fixed evaluation set of gold key--value instances $G$ and an available inventory $\widehat{\mathcal{K}}_{\mathrm{c}}$, let
\begin{equation}
\label{keyCoverage}
G_{\mathrm{cov}} =
\{(k,v)\in G:\psi(k)\in\widehat{\mathcal{K}}_{\mathrm{c}}\},
\qquad
C_{\mathrm{instance}} =
\frac{|G_{\mathrm{cov}}|}{|G|}.
\end{equation}
This instance-weighted definition is tied to a named corpus, split, inventory version, and train-only ranking; it does not estimate the unobservable universe of all possible clinical keys. We separately report conditional recall on supported instances,
\begin{equation}
R_{\mathrm{conditional}} = \frac{\mathrm{TP}}{|G_{\mathrm{cov}}|},
\end{equation}
and end-to-end recall on all gold instances,
\begin{equation}
R_{\mathrm{end\text{-}to\text{-}end}} = \frac{\mathrm{TP}}{|G|}.
\end{equation}
Because the QA system queries only inventory members, unsupported fields cannot enter its accepted output. Under the same pooled-micro matching universe, the recall identity
$R_{\mathrm{end\text{-}to\text{-}end}} =
C_{\mathrm{instance}}R_{\mathrm{conditional}}$ therefore holds. It is a recall decomposition, not a Precision or F1 decomposition. In the remainder of the
paper we refer to $C_{\mathrm{instance}}$ as instance coverage,
$R_{\mathrm{conditional}}$ as conditional recall, and
$R_{\mathrm{end\text{-}to\text{-}end}}$ as end-to-end recall.

\subsection{Training Method}
The Chinese coverage experiment uses BERT-based extractive QA models with
token-level start and end heads. Each Top-$N$ inventory is ranked from the
7,639 training reports only and paired with its corresponding model; all
conditions are evaluated on the same 849-report development set. During
inference, only canonical keys in the selected inventory are queried, and the
resulting surface-form keys and values are assembled into accepted key--value
pairs.

Long documents are handled separately in the two implementations. The primary
BERT-based QA experiment uses a 512-token question--context window and adapts
the context budget to the query length. It processes non-overlapping,
blank-line-delimited segments and aggregates predictions for each report--key
query; it neither uses a sliding overlap nor reconnects spans across segments.
The supplementary BERT--BiLSTM--CRF implementation instead uses overlapping
chunks, as specified in Appendix~\ref{app:chunking-config}. We report
pooled-micro Exact and BTM-3 precision, recall, and F1 over gold and accepted
predicted report--key units; Overlap is reported only as a separately labelled
diagnostic.

The conditional extraction comparison uses a separate protocol on
content-deduplicated data. Models are restricted to the same gold key universe
and evaluate conditional value extraction rather than the inventory-gated
end-to-end mechanism studied in the Chinese coverage experiment. The complete
split, filtering, seed, model, and decoding contracts are listed in
Appendix~\ref{app:reproducibility}.

\section{Results}
\subsection{Canonical Inventory and Iterative Expansion}
\begin{figure}[H]
    \centering

    \begin{minipage}[t]{0.48\textwidth}
        \centering
        \includegraphics[width=\linewidth,trim=8 6 8 6,clip]{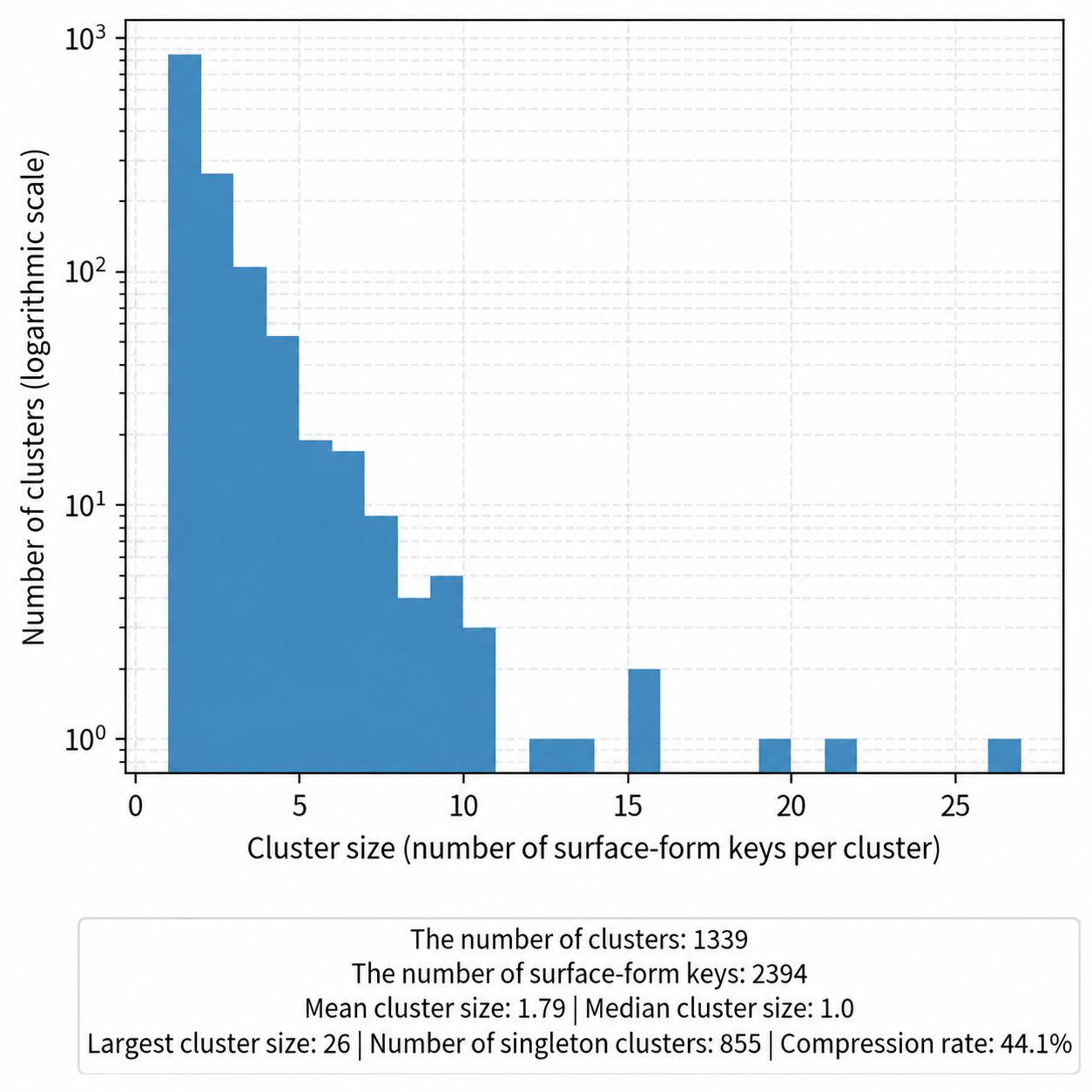}
        \subcaptiontext{(a) Number of clusters by cluster size}
    \end{minipage}
    \hfill
    \begin{minipage}[t]{0.48\textwidth}
        \centering
        \includegraphics[width=\linewidth,trim=8 6 8 6,clip]{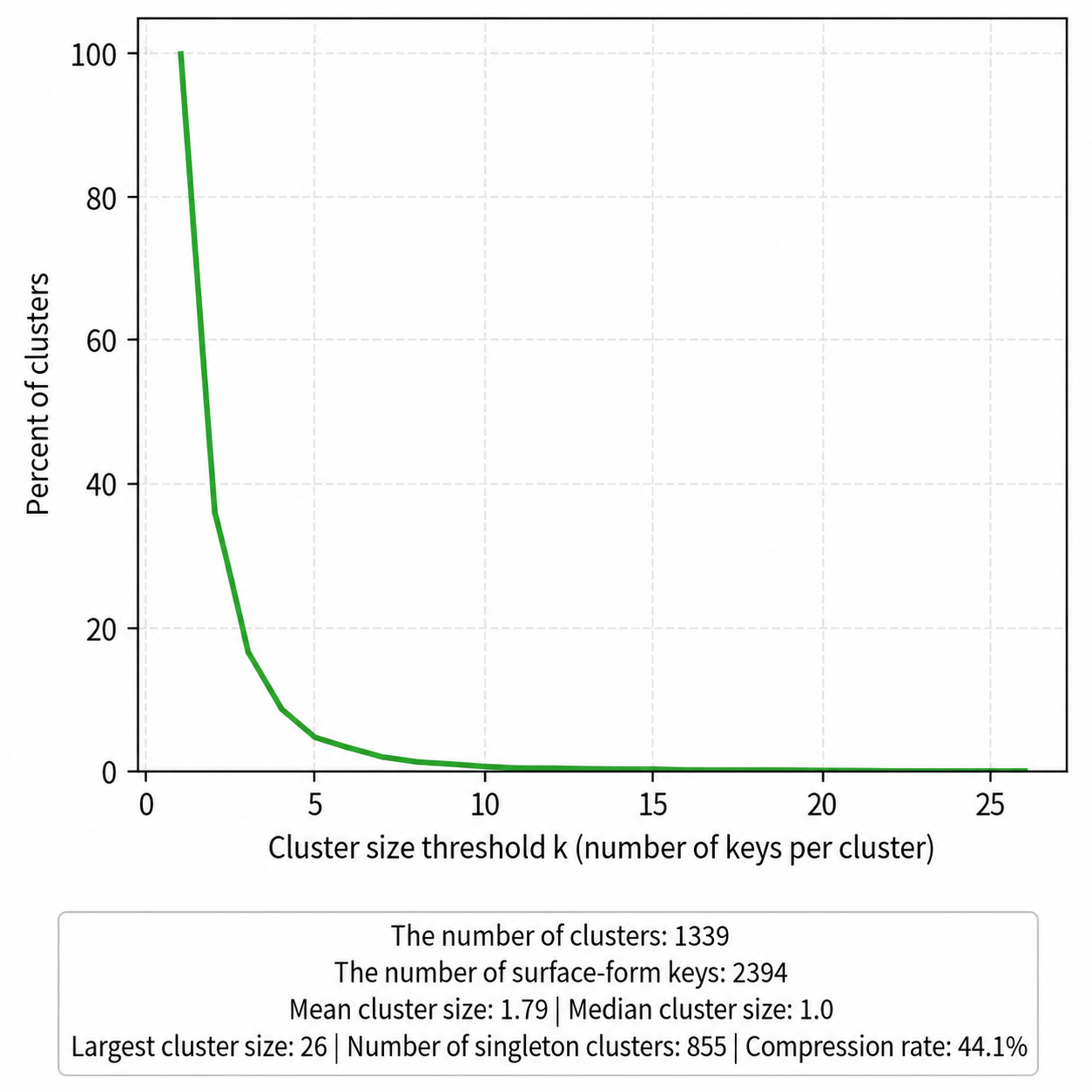}
        \subcaptiontext{(b) Complementary cumulative distribution function of cluster sizes}
    \end{minipage}

    \vspace{4pt}

    \begin{minipage}[t]{0.55\textwidth}
        \centering
        \includegraphics[width=\linewidth,trim=8 6 8 6,clip]{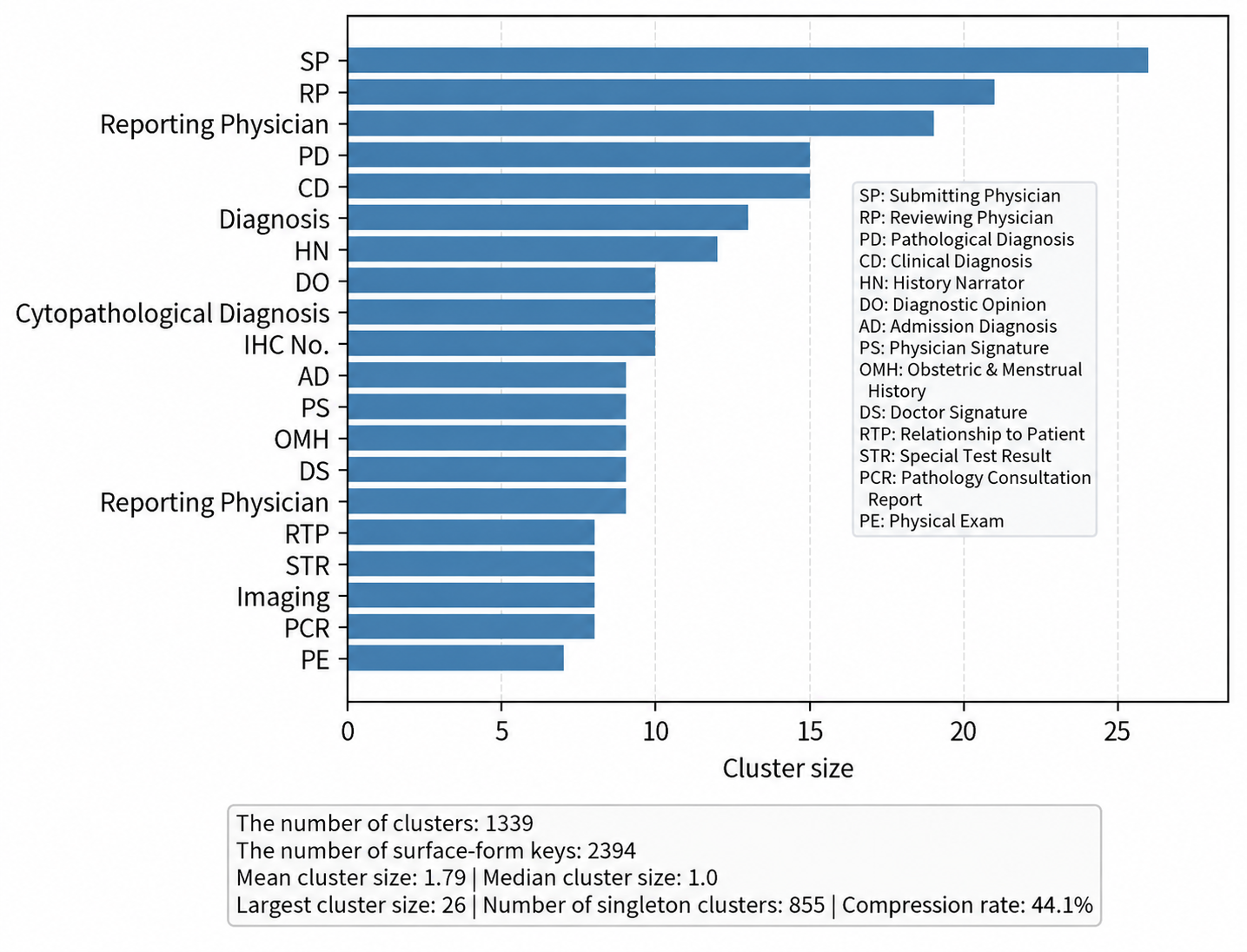}
        \subcaptiontext{(c) Sizes of the top 20 largest clusters}
    \end{minipage}

    \caption{Canonicalization clustering statistics of surface-form keys.}
    \label{fig:keyClustering}
\end{figure}

On the valid pre-deduplication annotation snapshot, canonicalization maps 2,394 observed surface-form keys to 1,339 canonical keys, a 44.1\% reduction in distinct key labels. This statistic describes lexical consolidation rather than compression of reports, key--value instances, or storage. Figure~\ref{fig:keyClustering} summarizes the resulting cluster-size distribution.

The mean cluster size is 1.79, the largest cluster contains 26 surface forms,
and 855 observed keys occur only once. These statistics illustrate both
substantial alias redundancy and a long tail of institution-specific
expressions. On the 536 surface keys shared by the iterative and one-shot
assignments, V-measure was 0.9331, indicating that incremental updating
preserved a similar grouping structure on the shared key set; additional
consistency metrics are reported in Appendix~\ref{app:clustering-consistency}.
This comparison did not test downstream extraction or human workload relative
to one-shot clustering.

\subsection{Chinese Coverage and End-to-End Recovery}
We evaluated seven train-derived
inventories (Top-10 through Top-100) on the same 849-report development set.
Each inventory was ranked from 7,639 training reports and used to constrain
the corresponding key-conditioned extraction model at inference. We report
instance coverage together with pooled-micro conditional and inventory-
constrained end-to-end precision, recall, and F1, separating the fields that
the inventory supports from the accuracy of extracting those fields.

\begin{table*}[t]
\centering
\small
\caption{Chinese instance coverage and pooled-micro Exact extraction across
train-derived inventories.}
\label{tab:chinese-coverage-exact}
\begin{tabular}{lrrrrrrr}
\toprule
Inventory & Coverage & \multicolumn{3}{c}{Conditional} &
\multicolumn{3}{c}{End-to-end} \\
\cmidrule(lr){3-5}\cmidrule(lr){6-8}
 &  & P & R & F1 & P & R & F1 \\
\midrule
Top-10 & 0.4019 & 0.8919 & 0.8794 & 0.8856 & 0.8919 & 0.3534 & 0.5062 \\
Top-20 & 0.5591 & 0.8647 & 0.8704 & 0.8675 & 0.8647 & 0.4866 & 0.6228 \\
Top-50 & 0.8233 & 0.8616 & 0.8653 & 0.8634 & 0.8616 & 0.7124 & 0.7799 \\
Top-80 & 0.9197 & 0.8490 & 0.8535 & 0.8513 & 0.8490 & 0.7850 & 0.8158 \\
Top-90 & 0.9339 & 0.8361 & 0.8571 & 0.8464 & 0.8361 & 0.8004 & 0.8179 \\
Top-95 & 0.9406 & 0.8365 & 0.8615 & 0.8488 & 0.8365 & 0.8103 & 0.8232 \\
Top-100 & 0.9465 & 0.8265 & 0.8497 & 0.8379 & 0.8265 & 0.8042 & 0.8152 \\
\bottomrule
\end{tabular}
\end{table*}

Table~\ref{tab:chinese-coverage-exact} shows that expanding the inventory
raises fixed-development instance coverage from 0.4019 at Top-10 to 0.9465
at Top-100. Over the same range, conditional Exact F1 decreases from 0.8856
to 0.8379, whereas inventory-constrained end-to-end recall rises from 0.3534
to 0.8042. End-to-end Exact F1 reaches 0.8232 at Top-95 and then decreases to
0.8152 at Top-100. Thus, in this corpus and protocol, inventory expansion
substantially broadens recoverable fields, while later additions exhibit
diminishing and non-monotone gains as conditional extraction becomes harder.

\subsection{Supplementary Experiments}
We compared BERT--BiLSTM--CRF with LoRA-adapted Qwen3
models~\citep{qwen3,lora} on a separate held-out set of 1,242
content-deduplicated reports. All models received the gold canonical key for
each query and extracted only the corresponding value. Qwen3-0.6B and Qwen3-1.7B achieved slightly
higher Exact F1, whereas BERT required substantially less inference time on the
same device. This comparison concerns conditional extraction, not open-world
coverage.

On the official DocILE~\citep{docile} split, expanding a train-derived inventory from Top-5
to Top-20 increases validation instance coverage from 45.40\% to 88.71\% and
the number of exactly recovered fields from 1,540 to 2,919, while conditional
Exact extraction decreases from 0.8271 to 0.8024. This provides external
evidence for the coverage--recovery mechanism in a public English document
domain, rather than evidence of transfer to English clinical reports. Under
controlled synthetic character corruption, the
BERT--BiLSTM--CRF model's Exact F1 decreases from 0.9060 at 0\% target character error rate (CER) to 0.1398 at 20\%; this is a
stress test, not a measured production OCR error distribution.

In an independent structured-representation study, adding structured fields to
raw text produced task-dependent effects: accuracy increased by 3.27 percentage
points for histologic grade and 3.52 points for treatment response, showed no
clear improvement for stage, recurrence, or primary site, and decreased by 5.54
points for 5-year survival. Because these fields were generated separately by
a private deployment of DeepSeek-V4-Flash~\citep{deepseek_v4}, this experiment
evaluates the downstream utility of an explicit structured representation rather
than the external validity of our extractor.

\section{Conclusion}
We study the clinically important problem of extracting header--content fields
from heterogeneous OCR clinical reports. We introduce an iterative
canonical-key-conditioned extraction pipeline that combines surface-key
mining, alias mapping, canonical inventory expansion, and incremental human
verification.

Inventory expansion improved end-to-end recall, but F1 gains diminished and
became non-monotone in the tail. The appropriate Top-$N$ operating point is
corpus- and protocol-specific rather than a universal constant.

\paragraph{Limitations.} The primary corpus reflects one oncology
patient-management setting. Because the archive lacks independent
pre-adjudication annotations, we report a
stratified retrospective revision audit rather than formal inter-annotator
agreement. OCR robustness was assessed with controlled synthetic corruption,
not production reports with paired human transcriptions. Accordingly, the
numerical operating points are corpus- and protocol-specific; within the
evaluated setting, canonical-key coverage remains an explicit constraint on
end-to-end field recovery.

\section*{Implementation Availability}
Sanitized implementation materials are available in the fixed release
\texttt{v0.1.0-camera-ready}:
\par\noindent\href{https://github.com/AI-Starfish-Research/Key-Coverage-Matters/tree/v0.1.0-camera-ready}{\nolinkurl{github.com/AI-Starfish-Research/Key-Coverage-Matters/tree/v0.1.0-camera-ready}}
\par
The repository contains code, configuration templates, synthetic examples,
and reviewed field-label-only canonical-inventory and alias snapshots.
Clinical-report images, OCR text, annotation exports, predictions, and
checkpoints are not publicly released.

\clearpage
\bibliography{mlhc26}

\clearpage
\appendix
\section{Supplementary Experiments and Implementation Details}

\subsection{Reproducibility protocol}
\label{app:ocr-config}
\paragraph{OCR preprocessing.}
Clinical-report images were processed page by page by a privately deployed
Baidu OCR V6 service~\citep{baidu_ocr}. Each request sent one page image as
base64-encoded content. The service used Chinese--English general text
recognition with page-direction detection and returned line- and
character-level text, confidence, and locations.

The response was normalized to line-level records containing recognized text,
rectangular location, and confidence; character records retained text, location,
and confidence when available. The model input concatenated line text in
service-returned order, while confidence and geometry remained
metadata; multi-page outputs were aggregated after extraction. This production
configuration is distinct from the synthetic corruption stress test in
Table~\ref{tab:ocr-noise}.

\paragraph{Long-document handling.}
\label{app:chunking-config}
Table~\ref{tab:chunking-config} distinguishes the query-conditioned budget
splitting used for Chinese coverage from the sliding-window and span-merging
path used for conditional extraction. The latter's 50-token overlap and span
reconnection do not apply to the Chinese coverage results.

\begin{table}[H]
\centering
\small
\caption{Long-document handling in the Chinese coverage experiment and the conditional extraction comparison.}
\label{tab:chunking-config}
\begin{tabular}{>{\RaggedRight\arraybackslash}p{0.25\textwidth}
                >{\RaggedRight\arraybackslash}p{0.33\textwidth}
                >{\RaggedRight\arraybackslash}p{0.33\textwidth}}
\toprule
Setting & Chinese coverage experiment & Conditional extraction comparison \\
\midrule
Input limit & 512 tokens for the question--context pair & 512 tokens \\
Nominal context/chunk size & \(\min(500,512-|q|-3)\) tokens for query \(q\) (nominal 500) & 500 tokens \\
Segmentation & Blank-line-delimited line segments; greedy budget splitting only when needed & Token sliding windows \\
Overlap / stride & None & 50-token overlap / 450-token stride \\
Truncation fallback & Re-split the affected candidate greedily at 80\% of its context budget; drop only a subchunk that still cannot be encoded & Each window is independently encoded within the 512-token limit \\
Cross-chunk aggregation & Per report--key query, select the maximum null score across chunks; no cross-chunk span concatenation & Convert local spans to report-level character offsets; remove duplicate \((\mathrm{type},s,e)\) spans; merge same-type overlapping or adjacent spans with gap \(\leq5\), then recover text from the full report \\
\bottomrule
\end{tabular}
\end{table}

\paragraph{Corpus snapshots.}
\label{app:corpus-flow}
Table~\ref{tab:corpus-flow} lists the data snapshots. Chinese coverage uses a
separate 8,488-report cohort.

\begin{table}[H]
\centering
\small
\caption{Corpus snapshots and evaluation sets.}
\label{tab:corpus-flow}
\begin{tabular}{>{\RaggedRight\arraybackslash}p{0.19\textwidth}
                >{\RaggedRight\arraybackslash}p{0.17\textwidth}
                >{\RaggedRight\arraybackslash}p{0.53\textwidth}}
\toprule
Snapshot & Count & Definition and use \\
\midrule
Collected reports & 16,144 & Pre-validity-screen census; acquisition and total-time reference only. \\
Valid annotation tasks & 15,991 & After 153 reports failed the validity screen; contains 132,798 KV pairs and 2,394 surface-form keys. \\
Deduplicated task pool & 14,587 & Exact full-OCR-text deduplication removed 1,404 tasks. \\
Seed-42 task partition & 13,128 / 1,459 & Training / held-out split of the deduplicated pool. \\
Held-out evaluation set & 1,458 & One split task had no evaluation record; used for conditional evaluation and timing. \\
Common evaluation subset & 1,242 & Held-out reports with valid gold spans; the 216 excluded reports lacked a valid span. \\
\bottomrule
\end{tabular}
\end{table}

\paragraph{Retrospective revision audit.}
To characterize annotation stability in the retained archive, we searched the
Label Studio PostgreSQL database, two SQLite snapshots, and 47 historical
exports. These sources did not recover demonstrably independent, blinded,
pre-adjudication annotation pairs, so formal inter-annotator agreement cannot
be computed. We therefore sampled 200 reports (40 from each of five report
categories) and compared the earliest and final retained versions of the same
annotation record. Table~\ref{tab:revision-audit} reports exact span or relation
F1 and the proportion of unchanged reports. It measures revision stability, not
independent inter-annotator agreement. Across the five equally sized strata,
KEY exact-span F1 ranged from 0.9228 to 0.9619 and VALUE exact-span F1 from
0.9167 to 0.9552.

\begin{table}[H]
\centering
\small
\caption{Revision stability between the earliest and final retained versions of the same annotation record (200 reports).}
\label{tab:revision-audit}
\begin{tabular}{lr}
\toprule
Outcome & Estimate \\
\midrule
KEY exact span-F1 & 0.9473 \\
VALUE exact span-F1 & 0.9424 \\
HOSPITAL exact span-F1 & 0.9776 \\
KEY$\rightarrow$VALUE relation-F1 & 0.9598 \\
Unchanged reports & 37.0\% \\
\bottomrule
\end{tabular}
\end{table}

\paragraph{Evaluation and inference settings.}
\label{app:reproducibility}
Table~\ref{tab:repro-protocols} contrasts the two primary protocols.

\begin{table}[H]
\centering
\small
\caption{Core evaluation protocols.}
\label{tab:repro-protocols}
\begin{tabular}{>{\RaggedRight\arraybackslash}p{0.23\textwidth}
                >{\RaggedRight\arraybackslash}p{0.34\textwidth}
                >{\RaggedRight\arraybackslash}p{0.34\textwidth}}
\toprule
Setting & Chinese coverage & Conditional extraction \\
\midrule
Snapshot and split & 8,488 reports; seed-42 7,639/849 train/development split & 14,587 deduplicated tasks; seed-42 13,128/1,459 train/held-out split \\
Duplicate handling & None & Exact full-OCR-text deduplication before splitting \\
Key condition & Training-ranked Top-$N$ inventories (10, 20, 50, 80, 90, 95, 100); canonical-key queries & Gold canonical keys; Qwen receives the gold key list, and CRF predictions are mapped to the same canonical set \\
Evaluation population & All 849 development reports at each Top-$N$ & 1,458 held-out reports; 1,242 common reports with valid gold spans \\
Metrics & Micro Exact and BTM-3 P/R/F1; Overlap diagnostic; conditional and end-to-end recovery separate & Micro Exact and BTM-3 P/R/F1 on the common subset; Overlap diagnostic \\
\bottomrule
\end{tabular}
\end{table}

\paragraph{Inference settings.} Chinese coverage uses archived checkpoints and
deterministic rescoring; its original training configuration was not retained.
Multilingual BERT extractive QA uses 512-token question--context pairs
(500-token context), batch size 1, query templates, and a 0.5 no-answer
threshold; each Top-$N$ inventory uses its matching model. The
BERT--BiLSTM--CRF baseline uses a three-layer BiLSTM (hidden size 384), CRF,
and BIOE labels; it trains for six epochs (batch size 8, learning rate
\(2\times10^{-5}\), weight decay 0.03, warm-up ratio 0.05, dropout 0.2, and
gradient clipping 1.0) and evaluates with batch size 16 and 500/50 chunk
aggregation. Qwen3-0.6B/1.7B use the LoRA SFT configuration in
Table~\ref{tab:lora-training-config} and greedy chat decoding (at most 2,048
new tokens; batch size 1; seed 42). Qwen3-32B uses one worked example and the
gold key list per record with greedy decoding (at most 2,048 new tokens; batch
size 8; seed 42).

\subsection{Supplementary Chinese coverage results}
Table~\ref{tab:chinese-coverage-edge3} provides the boundary-tolerant results
for the 849-report development set. All inventory-gate violation counts and
pooled-recall residuals are zero.

\begin{table}[H]
\centering
\small
\caption{Boundary-tolerant Chinese coverage results under the Chinese coverage experiment.}
\label{tab:chinese-coverage-edge3}
\begin{tabular}{lrrrrrrr}
\toprule
Inventory & Coverage & \multicolumn{3}{c}{Conditional} &
\multicolumn{3}{c}{End-to-end} \\
\cmidrule(lr){3-5}\cmidrule(lr){6-8}
 &  & P & R & F1 & P & R & F1 \\
\midrule
Top-10 & 0.4019 & 0.8919 & 0.8794 & 0.8856 & 0.8919 & 0.3534 & 0.5062 \\
Top-20 & 0.5591 & 0.8642 & 0.8700 & 0.8671 & 0.8642 & 0.4864 & 0.6224 \\
Top-50 & 0.8233 & 0.8613 & 0.8651 & 0.8632 & 0.8613 & 0.7122 & 0.7797 \\
Top-80 & 0.9197 & 0.8490 & 0.8535 & 0.8513 & 0.8490 & 0.7850 & 0.8158 \\
Top-90 & 0.9339 & 0.8358 & 0.8568 & 0.8462 & 0.8358 & 0.8002 & 0.8176 \\
Top-95 & 0.9406 & 0.8366 & 0.8616 & 0.8489 & 0.8366 & 0.8104 & 0.8233 \\
Top-100 & 0.9465 & 0.8265 & 0.8497 & 0.8379 & 0.8265 & 0.8042 & 0.8152 \\
\bottomrule
\end{tabular}
\end{table}

\subsection{Conditional extraction baselines}
Table~\ref{tab:conditional-model-comparison} compares three models on the
same 1,242-report subset. Gold canonical keys are supplied, so this is a
conditional extraction comparison rather than end-to-end recovery; timing and
memory were measured on one NVIDIA RTX 4090.

\begin{table}[H]
\centering
\small
\setlength{\tabcolsep}{4pt}
\begin{tabular}{lrrrrrr}
\toprule
Model & Exact F1 & BTM-3 F1 & Overlap F1 & Batch & Time (s) & Peak VRAM \\
\midrule
BERT--BiLSTM--CRF 0.2B & 0.90606 & 0.90643 & 0.97170 & 16 & 65.58 & 5,120 MiB \\
Qwen3-0.6B LoRA & 0.92023 & 0.92994 & 0.96098 & 1 & 6,155.24 & 5,892 MiB \\
Qwen3-1.7B LoRA & 0.92011 & 0.92889 & 0.96090 & 1 & 6,346.02 & 8,140 MiB \\
\bottomrule
\end{tabular}
\caption{Gold-key-conditioned extraction accuracy on the common evaluation subset of 1,242 reports. Time and peak VRAM denote full inference on the 1,458-report held-out evaluation set before valid-span filtering.}
\label{tab:conditional-model-comparison}
\end{table}

The Qwen baselines were fine-tuned on the same 90/10 snapshot with the
configuration in Table~\ref{tab:lora-training-config}. Both used
completion-only supervision, masking prompt tokens and preserving the target
completion when inputs exceeded the model-specific length limit.

\begin{table}[H]
\centering
\small
\setlength{\tabcolsep}{3.5pt}
\caption{LoRA SFT configuration for the Qwen3 baselines in the conditional extraction comparison. Effective batch size is per-device batch size multiplied by gradient accumulation.}
\label{tab:lora-training-config}
\begin{tabular}{>{\RaggedRight\arraybackslash}p{0.31\textwidth}
                >{\RaggedRight\arraybackslash}p{0.27\textwidth}
                >{\RaggedRight\arraybackslash}p{0.27\textwidth}}
\toprule
Setting & Qwen3-0.6B & Qwen3-1.7B \\
\midrule
LoRA rank / alpha / dropout & 16 / 32 / 0.05 & 16 / 32 / 0.05 \\
Target modules & \multicolumn{2}{l}{Attention Q/K/V/O and feed-forward gate/up/down projections} \\
Bias / task type & none / causal LM & none / causal LM \\
Base precision / quantization & bf16 / no 4-bit & bf16 / no 4-bit \\
Maximum sequence length & 4,096 & 2,048 \\
Epochs & 3 & 3 \\
Per-device batch / accumulation / effective batch & 1 / 16 / 16 & 1 / 16 / 16 \\
Optimizer / learning rate & AdamW / \(2\times10^{-4}\) & AdamW / \(2\times10^{-4}\) \\
Schedule / warm-up & cosine / 3\% & cosine / 3\% \\
Weight decay / max gradient norm & 0 / 1.0 & 0 / 1.0 \\
Gradient checkpointing & enabled & enabled \\
Loss / seed & completion only / 42 & completion only / 42 \\
\bottomrule
\end{tabular}
\end{table}

\paragraph{Qwen3-32B gold-key few-shot baseline.}
The Qwen3-32B constrained few-shot baseline obtains Exact, BTM-3, and Overlap
F1 scores of 0.65612, 0.66274, and 0.71156 on the common evaluation subset of
1,242 reports under gold-key conditioning, with six parse failures.

\subsection{Robustness and transfer analyses}
\paragraph{DocILE inventory expansion.}
On the official DocILE split (5,179 training and 500 validation documents),
training-ranked Top-5 and Top-20 inventories covered 45.40\% and 88.71\% of
the 4,101 validation fields, recovering 1,540 and 2,919 exact fields,
respectively. Conditional Exact/Overlap were 0.8271/0.8623 at Top-5 and
0.8024/0.8463 at Top-20.

\paragraph{Synthetic OCR-noise stress test.}
We applied synthetic character corruption to the 1,242-report common subset
(substitution/deletion/insertion probabilities 0.60/0.25/0.15) and evaluated
the BERT--BiLSTM--CRF conditional extractor. Table~\ref{tab:ocr-noise} is a
controlled stress test, not an estimate of production OCR error rates.

\begin{table}[H]
\centering
\small
\begin{tabular}{r@{\hspace{0.8em}}r@{\hspace{0.8em}}r@{\hspace{0.8em}}r@{\hspace{0.8em}}r}
\toprule
Target CER & Measured CER & Exact F1 & BTM-3 F1 & Overlap F1 \\
\midrule
0\% & 0.000\% & 0.9060 & 0.9062 & 0.9711 \\
2.5\% & 2.468\% & 0.6321 & 0.6325 & 0.7098 \\
5\% & 4.982\% & 0.4813 & 0.4819 & 0.5684 \\
10\% & 10.002\% & 0.3027 & 0.3028 & 0.4048 \\
15\% & 14.998\% & 0.2063 & 0.2063 & 0.3037 \\
20\% & 20.011\% & 0.1398 & 0.1400 & 0.2400 \\
\bottomrule
\end{tabular}
\caption{Performance under synthetic character corruption.}
\label{tab:ocr-noise}
\end{table}

\paragraph{Iterative versus one-shot clustering consistency.}
\label{app:clustering-consistency}
On the 536 surface keys shared by iterative and one-shot canonicalization,
homogeneity is 0.9326, completeness is 0.9335, ARI is 0.3850, and directional
purities are 0.8078 and 0.7873 (V-measure $0.9331$ is reported in the main
text). These measurements establish clustering consistency only.

\subsection{Independent downstream utility study}
Table~\ref{tab:tcga-utility} reports task-specific estimates from the
independent utility study using 10,000 case-level paired bootstrap resamples
(seed 42). TEXT is the text-only baseline; TEXT+STRUCT augments it with the
structured representation.

\begin{table}[H]
\centering
\scriptsize
\resizebox{\textwidth}{!}{%
\begin{tabular}{lrrrrrl}
\toprule
Task & $n$ & TEXT & STRUCT & TEXT+STRUCT & $\Delta$ vs. TEXT & 95\% paired bootstrap CI \\
\midrule
Stage & 578 & 0.7111 & 0.2976 & 0.7128 & +0.17 pp & [-2.25, +2.60] pp \\
Histologic grade & 672 & 0.8378 & 0.6057 & 0.8705 & +3.27 pp & [+0.60, +5.95] pp \\
Treatment response & 825 & 0.3042 & 0.1479 & 0.3394 & +3.52 pp & [+1.45, +5.58] pp \\
Tumor recurrence & 1,481 & 0.5078 & 0.4031 & 0.5219 & +1.42 pp & [-0.07, +2.84] pp \\
5-year survival & 578 & 0.1211 & 0.0121 & 0.0657 & -5.54 pp & [-7.79, -3.46] pp \\
Primary site & 1,581 & 0.9690 & 0.2460 & 0.9677 & -0.13 pp & [-0.63, +0.38] pp \\
\bottomrule
\end{tabular}
}
\caption{Accuracy differences in the independent TCGA
structured-representation utility study. TEXT, STRUCT, and TEXT+STRUCT are
the three paired input arms; the last two columns report TEXT+STRUCT minus
TEXT.}
\label{tab:tcga-utility}
\end{table}

\end{CJK}
\end{document}